\documentclass[letterpaper, 10 pt, conference]{ieeeconf}
\IEEEoverridecommandlockouts 
\usepackage{cite}
\usepackage{amsmath,amssymb,amsfonts}
\usepackage{algorithmic}
\usepackage{graphicx}
\usepackage{textcomp}
\let\labelindent\relax
\usepackage[inline]{enumitem}
\usepackage{xcolor}
\usepackage{caption}
\usepackage{graphicx}
\usepackage{algorithm}  
\makeatletter
\newcommand\fs@betterruled{%
  \def\@fs@cfont{\bfseries}\let\@fs@capt\floatc@ruled
  \def\@fs@pre{\vspace*{5pt}\hrule height.8pt depth0pt \kern2pt}%
\def\@fs@post{\kern2pt\hrule\relax}%
  \def\@fs@mid{\kern2pt\hrule\kern2pt}%
  \let\@fs@iftopcapt\iftrue}
\floatstyle{betterruled}
\restylefloat{algorithm}
\makeatother

\usepackage{subfigure}
\usepackage[T1]{fontenc}
\usepackage{aecompl}
\usepackage{physics}

\usepackage{diagbox,slashbox}
\usepackage{hyperref}
\usepackage{verbatim}
\usepackage{dsfont}
\def\BibTeX{{\rm B\kern-.05em{\sc i\kern-.025em b}\kern-.08em
    T\kern-.1667em\lower.7ex\hbox{E}\kern-.125emX}}

\newenvironment{myitemize}{\begin{list}{$\bullet$}
{\setlength{\topsep}{1mm}
\setlength{\itemsep}{0.25mm}
\setlength{\parsep}{0.25mm}
\setlength{\itemindent}{0mm}
\setlength{\partopsep}{0mm}
\setlength{\labelwidth}{15mm}
\setlength{\leftmargin}{4mm}}}{\end{list}}

\newtheorem{definition}{Definition}

\begin{document}


\title{\LARGE \bf Kinematics-aware Trajectory Generation and Prediction with \\ Latent Stochastic Differential Modeling\\
}


\author{Ruochen Jiao$^{*1}$, Yixuan Wang$^{*1}$, Xiangguo Liu$^{1}$, Simon Sinong Zhan$^{1}$, Chao Huang$^{2}$, Qi Zhu$^{1}$
\thanks{$^{*}$Contribute equally to this work.}
\thanks{$^{1}$Ruochen Jiao, Yixuan Wang, Xiangguo Liu, Simon Sinong Zhan, and Qi Zhu are with the Department of Electrical and Computer Engineering, Northwestern University, IL, USA.}%
\thanks{$^{2}$Chao Huang is with the School of Electronics and Computer Science, the University of Southampton, UK.}%
}



\maketitle

\begin{abstract}
 Trajectory generation and trajectory prediction are two critical tasks in autonomous driving, which generate various trajectories for testing during development and predict the trajectories of surrounding vehicles during operation, respectively. In recent years, emerging data-driven deep learning-based methods have shown great promise for these two tasks in learning various traffic scenarios and improving average performance without assuming physical models.
 However, it remains a challenging problem for these methods to ensure that the generated/predicted trajectories are physically realistic. This challenge arises because learning-based approaches often function as opaque black boxes and do not adhere to physical laws. Conversely, existing model-based methods provide physically feasible results but are constrained by predefined model structures, limiting their capabilities to address complex scenarios. To address the limitations of these two types of approaches, we propose a new method that 
integrates kinematic knowledge into neural stochastic differential equations (SDE) and designs a variational autoencoder based on this latent kinematics-aware SDE (\emph{LK-SDE}) to generate vehicle motions. Experimental results demonstrate that our method significantly outperforms both model-based and learning-based baselines in producing physically realistic and precisely controllable vehicle trajectories. Additionally, it performs well in predicting unobservable physical variables in the latent space.
\end{abstract}


\section{Introduction}
\begin{figure}[htbp]
    \centering
    \includegraphics[width=1\columnwidth]{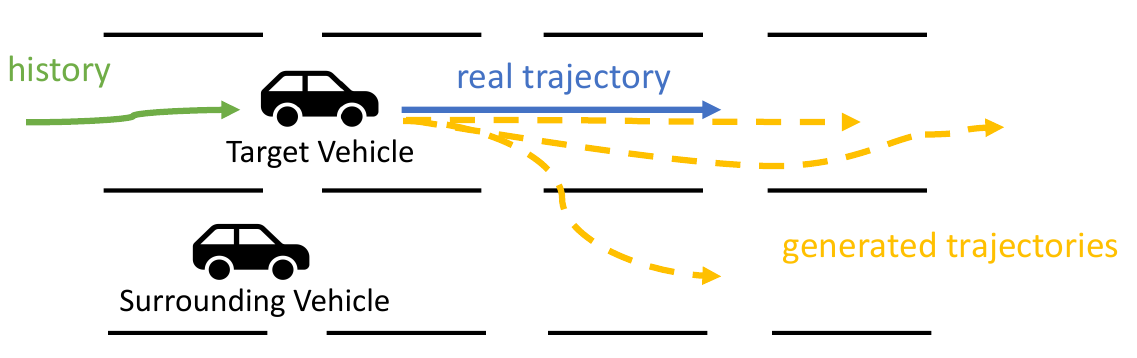}
    \caption{Generating diverse, physically realistic, and controllable trajectories is important for critical scenario augmentation and vehicle motion prediction.}
    \label{fig:intro_fig}
\end{figure}

Trajectory prediction and generation are two critical tasks for autonomous vehicles. First, as a key component in the autonomous driving pipeline, the \emph{trajectory prediction} module predicts the future trajectories of surrounding vehicles based on their recent trajectory histories (as observed by the ego vehicle) and the map information. 
The prediction result provides a safe operation space for downstream behavioral-level decision-making and motion planning~\cite{hu2023planning,liu2023safety,liu2022physics} and is critical for vehicle safety during operation. 
Then, given the long-tailed nature of real traffic scenarios, the important \emph{trajectory generation} task generates additional synthetic but realistic trajectories to augment the trajectory dataset collected in operation,  
for testing and optimizing the downstream planning module~\cite{ding2023survey,cai2022survey}. For instance, as shown in Fig.~\ref{fig:intro_fig}, we can convert a common and simple scenario to various challenging scenarios in the simulation by generating diverse trajectories and using them to test the reaction of the autonomous vehicle. 

In the literature, most works focus on improving the average accuracy for trajectory prediction and enumerating various scenarios for trajectory generation. 
However, the predicted/generated trajectories may not be realistic or even physically feasible in real traffic scenarios, and could lead to inferior training of the planning module and reduced capability of addressing safety-critical scenarios in practice. It is thus very important to ensure that the predicted/generated trajectories not only reflect the rich contextual factors, including other vehicles and HD maps, but also \emph{conform to traffic rules and fundamental laws of physics}. 

It is, however, challenging for the trajectory predictor and generator to learn the representation of high-dimensional context while aligning with physical constraints.
Traditional model-based techniques leverage simplified kinematic models, e.g., bicycle models, for trajectory generation. However, these methods often provide a very rough estimation of the vehicle's real motion and oversimplify the environment model without considering the surroundings, leading to coarse-grained and maladaptive generated trajectories in complex
traffic scenarios. On the other hand, learning-based generative models may effectively perceive the high-dimensional environment by learning from trajectory data and map context, as shown in recent advancements in trajectory prediction~\cite{liu2021multimodal,liang2020learning,ye2021tpcn,yuan2021agentformer} and generation~\cite{yin2021diverse,ding2020cmts,ding2023causalaf,rempe2022generating}. However, due to the lack of consideration of physical models, these generative techniques have little control over finer-grained vehicle-level kinematics and may produce physically unrealistic trajectories in real traffic scenarios.

In this paper, we aim to address the fundamental challenge in \textbf{simultaneously considering the complex surrounding environment and the vehicle physical model} for trajectory generation and prediction. The major difficulty is from the conflict between high-dimensional environment (captured by deep learning) and low-dimensional physics. To address this, we utilize \textbf{latent kinematics-aware neural stochastic differential equations \emph{(LK-SDE)}} to bridge the model-based kinematics and high-dimensional representations for road contexts in the latent space of a variational autoencoder (VAE)~\cite{kingma2013auto}.
Specifically, we first extract the graph convolutional network (GCN) representation from the HD maps and  environment, which convey the information of the environment contexts. Together with a physical model, the GCN representation is then used to train a neural SDE. Such kinematics-aware SDE is optimized and calculated in the latent space of the VAE, providing additional critical information for prediction and precise control for decoding the trajectory generation. It is worth noting that this architecture can be added to various scenario augmentation and trajectory prediction frameworks to enhance physical realism and controllability. 

Our contributions in this work are summarized as follows:

\begin{myitemize}

\item We design a trajectory generator based on kinematics-guided SDE in the latent space, which effectively \emph{embeds physical constraints into deep learning models}. 

\item For the trajectory generation task, compared with pure deep learning-based and model-based methods, our method can \emph{generate more physically realistic and controllable augmented trajectories} by manipulating the kinematics-aware latent space. 

\item For the trajectory prediction task, our method can jointly \emph{predict realistic trajectories and important kinematic states} that are difficult to directly observe. 

\item Extensive experiments show that our method outperforms baseline methods across various metrics. These improvements will benefit the augmentation of safety-critical scenarios and the prediction of future motions.


\end{myitemize}

This paper is organized as follows. Section~\ref{sec:bg} introduces the background of trajectory generation and prediction and neural SDE. Section~\ref{sec:methods} presents the design of our \emph{LK-SDE} based VAE for trajectory generation and prediction. Experiment results are shown in Section~\ref{sec:exp} and Section~\ref{sec:con} concludes the paper. 

\section{Background}
\label{sec:bg}
\subsection{Trajectory Generation and Prediction}
Trajectory generation or augmentation plays an important role in evaluating and optimizing the decision-making module in autonomous driving. Many works use various deep learning methods to generate trajectories and scenarios. For instance, the work in~\cite{ding2020cmts} proposes a VAE-conditioned method to bridge safe- and collision-driving data to generate the whole risky scenario, but it cannot control agent-level trajectories. The work in~\cite{yin2021diverse} designs a GAN-based method -- RouteGAN to generate diverse trajectories for every single agent, and the trajectory is controlled by a style variable. Some recent approaches~\cite{ding2023causalaf,rempe2022generating,jiao2022tae} further utilize domain knowledge such as causal relations and traffic priors to generate useful traffic scenarios. However, these methods mainly focus on scenario-level generation. For trajectory generation of a single vehicle, these works still rely on pure deep learning methods or model-based methods. The latent spaces of these generative methods are not well modeled or explained for a single vehicle, especially at the kinematics or dynamics level. The models only have coarse and limited control over the generated trajectories, which often lead to physically unrealistic and uncontrollable trajectories.  

Recent works applied advanced deep learning techniques to learn the representations of agents' trajectories and road contexts. Graph neural networks~\cite{liang2020learning,zhao2020tnt}, transformer~\cite{liu2021multimodal,yuan2021agentformer,nayakanti2023wayformer}, and diffusion models~\cite{jiang2023motiondiffuser} are used to extract context features. The approach in~\cite{cui2020deep} adds a bicycle model after the neural feature extractor to decode the trajectories but their pure model-based decoder still suffers from the oversimplification of the vehicle motion. In this work, we aim to combine the powerful representations from deep learning models with kinematics knowledge in the latent space.

\subsection{Neural Stochastic Differential Equation}
The physical dynamics of many real-world systems can be modeled as a discrete-time SDE~\cite{kloeden1992stochastic} with the consideration of uncertainty and stochasticity.

\begin{definition}
    \textbf{(Stochastic Differential Equation (SDE))}: An SDE is a differential equation that contains stochastic processes, which can be expressed as 
    \begin{equation}\label{eq:physical_env_sde}
    \vb{s}_{t+1} = f(\vb{s}_t) + g(\vb{s}_t) \Delta W_t,
    \end{equation}
    where $\vb{s}_t \in \mathbb{R}^{n}$ is the system state, $f: \mathbb{R}^n \rightarrow \mathbb{R}^n$ denotes a drift function, and $g: \mathbb{R}^n \rightarrow \mathbb{R}^{n\times d}$ represents a diffusion function. $\Delta W_t = W(t+1) - W(t)$, where $ W(t) \in \mathbb{R}^d$ is the Brownian Motion (also known as Wiener Process)~\cite{cinlar2013introduction} for encoding the stochasticity in the systems. The Brownian Motion has the following properties: 
    \begin{itemize}
        \item $W(0) = 0$,
        \item $W(t)$ is almost surely continuous, 
        \item $W(t_1) - W(t_0) \sim \mathcal{N}(0, t_1-t_0)$, where $\mathcal{N}(0, t_1-t_0)$ is the Gaussian distribution with $0$ mean and $t_1-t_0$ variance. 
    \end{itemize}
\end{definition}

\begin{definition}
    \textbf{(Neural SDE)}: A neural SDE is an SDE with its drift function $f(\vb{s}_t)$ and diffusion function $g(\vb{s}_t)$ expressed and parameterized by deep neural networks, e.g., $f_{\theta_0}(\vb{s}_t), g_{\theta_1}(\vb{s}_t)$~\cite{li2020scalable}: 
    \begin{equation}\label{eq:nsde}
    \hat{\vb{s}}_{t+1} = f_{\theta_0}(\hat{\vb{s}}_t) + g_{\theta_1}(\hat{\vb{s}}_t) \Delta W_t.
    \end{equation}
\end{definition}

To learn such neural network representations, a typical way is to sample the real physical environment as Eq.~\eqref{eq:physical_env_sde} and neural SDE as Eq.~\eqref{eq:nsde} to generate the trajectory data $\tau = \{\vb{s}_0, \vb{s}_1, \cdots, \vb{s}_T\}$ and $\hat{\tau} = \{\hat{\vb{s}}_0, \hat{\vb{s}}_1, \cdots, \hat{\vb{s}}_T\}$, respectively. Then fit the real-world trajectory $\tau$ to the neural networks by reducing the following loss function: 
\begin{displaymath}
\min_{\theta_0, \theta_1}\mathcal{L}(\tau, \hat{\tau}) = \min_{\theta_0, \theta_1} - \sum_{t=0}^{T}  \log \left( P\left(\vb{s}_t ~|~ \mathcal{N}(\hat{\vb{s}}_t, g_{\theta_1}(\hat{\vb{s}}_t)\right)\right),
\end{displaymath}
where $\mathcal{L}$ is the maximum likelihood loss, $T$ is the time length, and $P\left(\vb{s}_t ~|~ \mathcal{N}(\hat{\vb{s}}_t, g_{\theta_1}(\hat{\vb{s}}_t)\right)$ is the likelihood probability of the observed $\vb{s}_t$ under the normal distribution $\mathcal{N}(\hat{\vb{s}}_t, g_{\theta_1}(\hat{\vb{s}}_t))$ of the neural SDE at time $t$. 
\section{Our LK-SDE Methods}
\label{sec:methods}
\begin{figure*}[htbp]
    \centering
    \includegraphics[width=1.9\columnwidth]{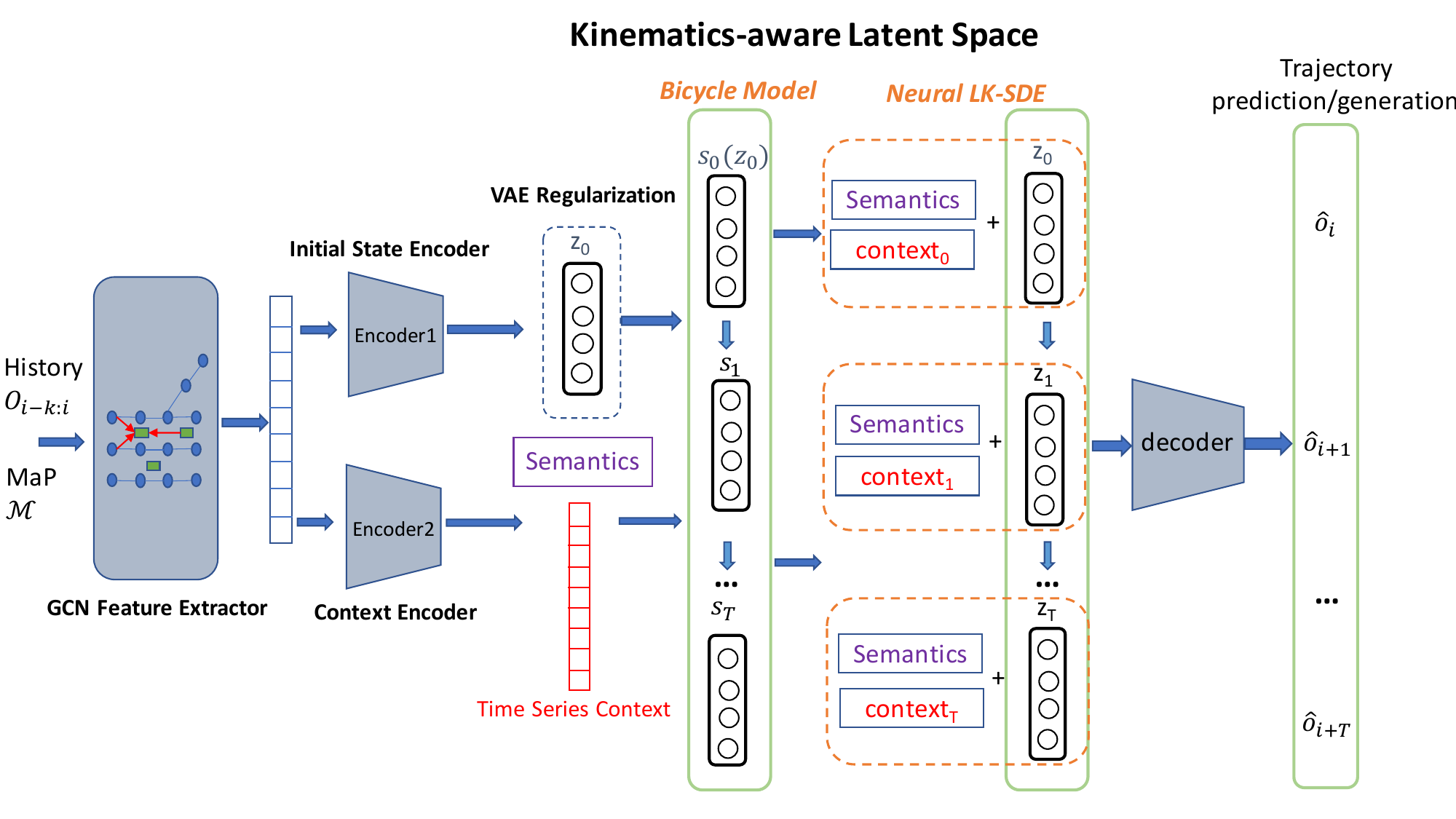}
    \caption{The overall design of our kinematics-aware trajectory generator. The GCN and two individual encoders consume the driving historical trajectories $o_{i-k:i}$ and map information $\mathcal{M}$. They extract and generate the latent initial state $z_0$, global semantic feature $\vb{sem}$, and context feature per step $\vb{ctx}_t$ for the latent space. Within the latent space, we learn a kinematics-aware neural SDE guided by a physical bicycle model and then decode the latent vectors $\vb{z}_{i} (i=0, \cdots, T)$ to the output vehicle motion trajectory $\hat{o}_{i: i + T}$. Our neural \emph{LK-SDE} is guided by the kinematic bicycle model during training to learn physical knowledge for physically feasible and controllable trajectory generation and prediction.}
    \label{fig:pipeline}
\end{figure*}

\subsection{Overall Design}
\noindent
\textbf{Problem Set Up:} We study the trajectory prediction and generation tasks that aim to predict and generate future two-dimensional trajectory waypoints given the driving histories of the driving cars as well as the map information. We are given a dataset of trajectory observations and map information as $(o_{i}, \mathcal{M}) (i = 0, \cdots, N)$. Assuming at timestamp $i$, the input for our entire model is a $k$-length two-dimensional history trajectories $o_{i-k: i}$ of vehicles (including both target and surrounding vehicles) and the $\mathcal{M}$ graph of HD maps. The output of the model is the generated or predicted future trajectories $\hat{o}_{i:i+T}$, where $T$ is the prediction and generation horizon in the future. 

\smallskip
\noindent
\textbf{Overview:} Our method is a learning-based VAE approach with kinematics-aware latent space guided/constrained by the kinematics knowledge from a bicycle model. 
The overall architecture of the proposed method is shown in Fig.~\ref{fig:pipeline}. We first feed 
the trajectories of vehicles $o_{i-k: i}$ and map contexts $\mathcal{M}$ into a GCN-based feature extractor to learn representations of HD maps and interactions between vehicles. The extracted features are further processed by two encoders -- one encoder converts the representations to a four-dimensional latent initial state, and 
the other encoder will generate a global semantic vector and finer-grained time-series contexts for future steps in the latent space. 
As a model-based approach, our neural \emph{LK-SDE} takes the semantic vector, corresponding context, and the initial state from learning as inputs and rolls out the latent states step by step via learned latent dynamics.   The neural \emph{LK-SDE} states are guided to be close to the bicycle-model latent states and thus we embed the kinematics knowledge from the bicycle model into the latent dynamics. Finally, a simple fully-connected neural network will decode the latent states into vehicle trajectory space. The bicycle model guided latent dynamics in our VAE-like approach contribute to more physically feasible and controllable trajectory learning due to the latent space constraints, compared to existing learning-based approaches.  

The algorithm of our approach is shown in Algorithm~\ref{alg:pipeline}. We introduce the details of each submodule in the following. 

\begin{algorithm}[!t]
\caption{\raggedright Optimization Pipelines}
\label{alg:pipeline}
\begin{algorithmic}[1]
\STATE\textbf{Initialize:} feature extractor $G$, initial state encoder $E_s$, context encoder $E_c$, decoder $D$, \emph{LK-SDE}($f_{\theta_0}$,$g_{\theta_1}$), and bicycle-model SDE($h(\cdot, \pi)$,$g_{\theta_1}$)

\STATE\textbf{Input:} past trajectories $o_{i-k:i}$ and map graph $\mathcal{M}$.

\FOR{each batch}
\STATE Let features $x$ = $G(o_{i-k :i},\mathcal{M})$.
\STATE Let initial kinematic vectors $\vb{z}_0$ = $\vb{s}_0$ = $E_s(x)$.
\STATE Let global semantics and time series contexts\\ $\vb{sem}$ , $\vb{ctx}_{1,2,...,T}$ = $E_c(x)$.
\STATE Update the $G$ and $E_s$ by the regularization loss $L_{reg}$ in Eq.~\eqref{eq:KL_reg}.
\FOR{$t$ in \textbf{range}($T$)}
\STATE  \emph{LK-SDE} computes $\vb{z}_{t+1} = f_{\theta_0}(\vb{z}_t, \vb{ctx}_t, \vb{sem}) + g_{\theta_1}(\vb{z}_t)\Delta W_t$.
\STATE Bicycle model SDE computes $\vb{s}_{t+1} = h(\vb{s}_t, \pi) + g_{\theta_1}(\vb{s}_t)\Delta W_t$.
\ENDFOR
\STATE Update the \emph{LK-SDE} ($f_{\theta_0}, g_{\theta_1}$) by the kinematic loss $L_{kin}$ in Eq.~\eqref{eq:kinematic_loss}.
\STATE The decoder projects latent vectors into trajectory space $\hat{o}_{i: i+T} = D(\vb{z}_0, \vb{z}_1, \cdots, \vb{z}_T)$.
\STATE Update the $G$, $E_s$, $E_c$, \emph{LK-SDE} ($f_{\theta_0}$,$g_{\theta_1}$) and $\pi$ by the prediction loss $L_{pred}$ in Eq.~\eqref{eq:l1-smooth}.
\ENDFOR
\end{algorithmic}
\end{algorithm}


\subsection{Encoders for Context Extraction and Embedding}
Similar to \cite{liang2020learning}, we use a one-dimensional convolutional network to model history trajectories, for its effectiveness in extracting multi-scale features
and efficiency in parallel computing. Multiple graph neural networks $G$ are utilized to learn the interactions among agents and lane nodes. After fusing the features $x$ of graphs and agents, we have two encoders $(E_s, E_c)$ to generate contexts and the initial state $\vb{z}_0 (\vb{s}_0)$ for our \emph{LK-SDE}, as shown in Fig.~\ref{fig:pipeline}. A ResNet~\cite{he2016deep} is applied in the context encoder to further generate global semantics $\vb{sem}$ and time-series local contexts $\vb{ctx}_{1: T}$ for future time steps $i + (1, \cdots, T)$, which is expressed as:
\begin{displaymath}
    \begin{aligned}
    x = G(o_{i-k :i},\mathcal{M}), \quad \vb{z}_0 = E_s(x), \\ \vb{sem}, \vb{ctx}_{1: T} = E_c(x).
    \end{aligned}
\end{displaymath}
Therefore, for each time step in the prediction horizon $t \in [1, T]$, we will have eight-dimensional local contexts $\vb{ctx}_t$ and four-dimensional global semantics $\vb{sem}$. The latent initial state $\vb{z}_0$ by the initial state encoder in Fig.~\ref{fig:pipeline} serves as the starting point for our neural \emph{LK-SDE} and bicycle model within the latent space. $\vb{z}_0$ is regularized to follow a Gaussian distribution by minimizing the Kullback–Leibler divergence as shown in Eq.~\eqref{eq:KL_reg}:
\begin{equation}
L_{reg} = \mathcal{KL}(q(\vb{z}_0|x)||p(\vb{z}_0)),
\label{eq:KL_reg}
\end{equation}
where $x$ is the input features, $p(\vb{z}_0)$ represents the targeted Gaussian distribution of the latent initial state, and $q$ represents the posterior distribution from the initial state encoder. The initial latent states $\vb{z}_0$ will be the input for the following \emph{LK-SDE}, and the global semantics $\vb{sem}$ as well as local context per step $\vb{ctx}_t$ will be the condition.

\subsection{Latent Kinematics-Aware SDE Modeling}
\begin{figure}[htbp]
    \centering
    \includegraphics[width=0.98\linewidth]{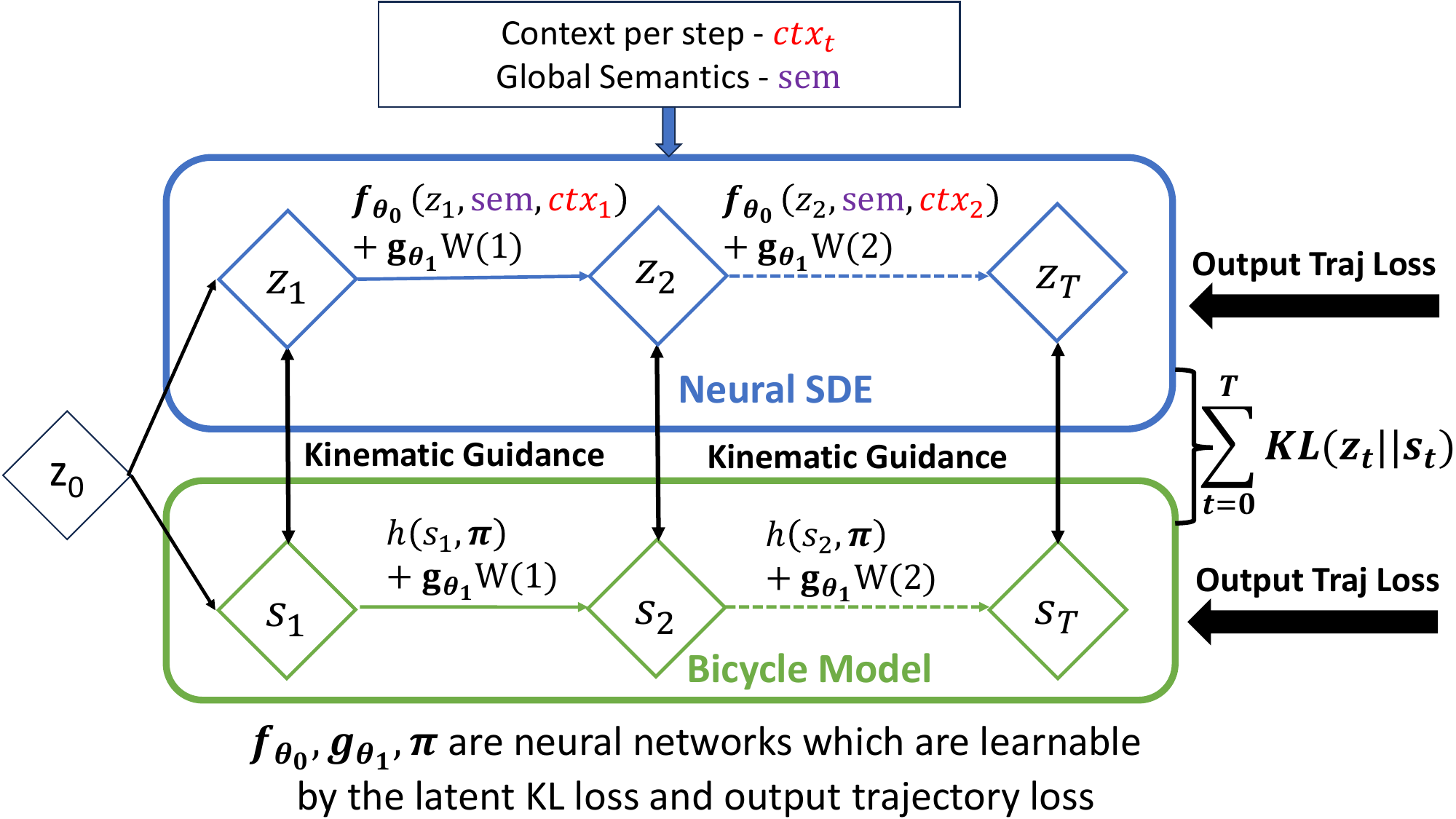}
    \caption{During the training, in the latent space, the bicycle model SDE guides our neural \emph{LK-SDE} to follow the kinematics by minimizing the KL divergence between the solutions of two SDEs. The kinematic loss function $L_{kin}$ is explained in Eq.~\eqref{eq:kinematic_loss}. $f_{\theta_0}$, $g_{\theta_1}$ are the neural networks for \emph{LK-SDE} that are optimized from the bicycle model in Eq.~\eqref{eq:kinematic_loss} and output loss function in Eq.~\eqref{eq:l1-smooth}, while $\pi$ in the bicycle model is optimized by the output loss function in Eq.~\eqref{eq:l1-smooth}.}  
    \label{fig:kinematics_guidance}
\end{figure}

The generation and prediction of vehicle trajectories rely on understanding the inherent dynamics and physical laws governing these trajectories. Consequently, this compels us to focus on acquiring a latent space that is attuned to the kinematics. More specifically, within the scope of this work, our objective is to acquire a \emph{LK-SDE} for the purposes of motion prediction and generation. This involves not only optimizing the loss function based on the ground truth label but also incorporating supervision from an explicit bicycle model~\cite{polack2017kinematic}, to serve as a constraint and guidance for the \emph{LK-SDE}. Specifically, the kinematics-aware latent space modeling involves two SDEs during training -- one is a learnable bicycle-model-based SDE to generate the kinematics based on the most recent states, and the other is the neural \emph{LK-SDE} that we optimized to both learn from the loss function of the output and learn the kinematics from the bicycle model SDE. The detailed kinematics-guided dual SDE learning process is illustrated in Fig.~\ref{fig:kinematics_guidance}. 

\subsubsection{Learnable Bicycle Model SDE}
The bicycle-model-based SDE is designed to infuse a comprehensive understanding of physics into our neural \emph{LK-SDE}. Over time, the latent state evolves according to this SDE, enabling us to capture a series of kinematics in the latent space of the VAE. This latent trajectory is subsequently decoded to produce the final task output, thereby increasing the likelihood of adhering to the specified physical constraints.

\begin{figure}[htbp]
    \centering
    \includegraphics[width=\columnwidth]{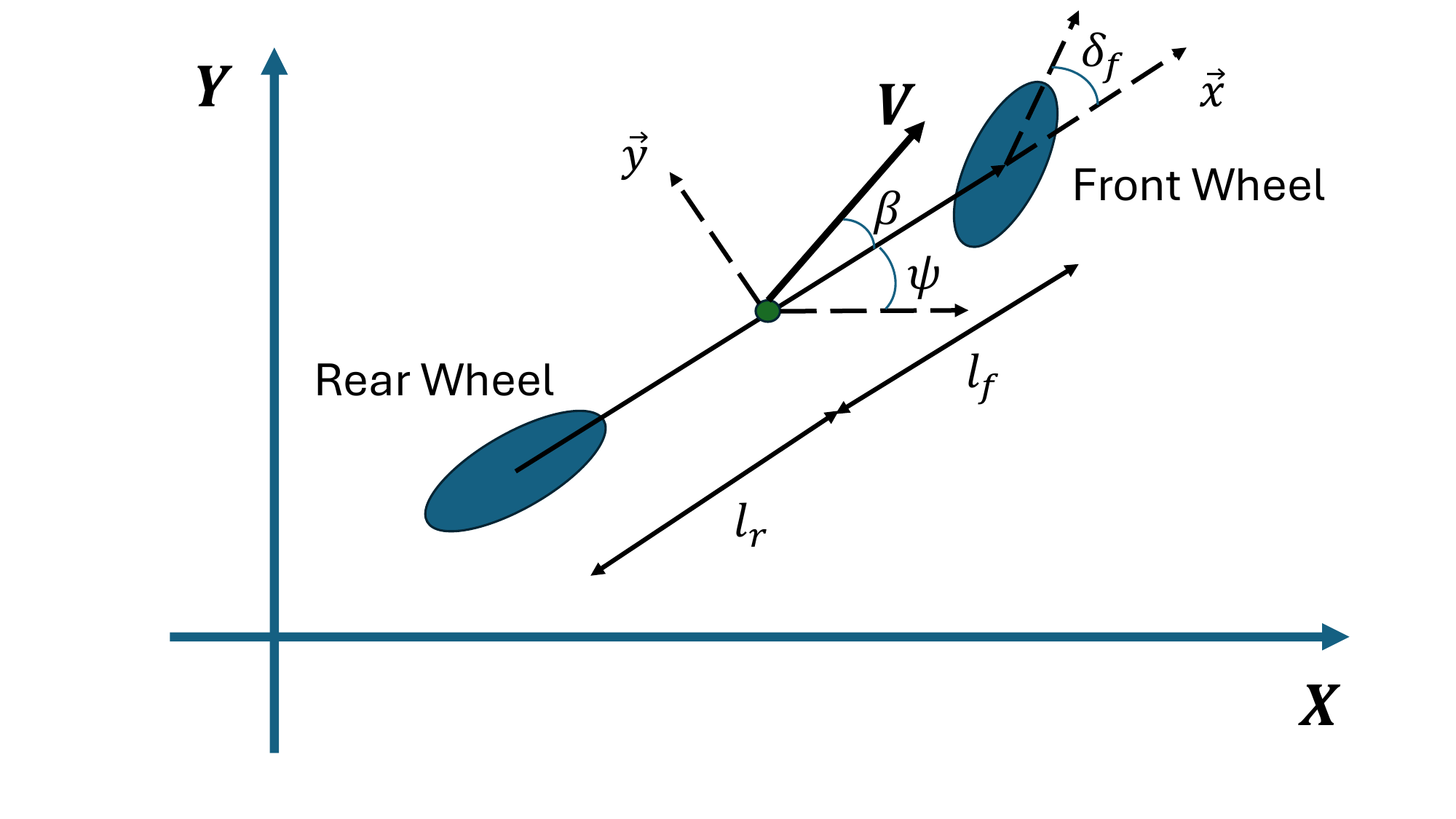}
    \caption{Illustration of the bicycle model\cite{polack2017kinematic}.}
    \label{fig:bicycle_model}
\end{figure}

We assume that the bicycle model SDE has $h(\vb{s}_t, \pi)$ as its drift function, where we differentiate by using $\vb{s}_t$, rather than $\vb{z}_t$ in \emph{LK-SDE}, $g_{\theta_1}(\vb{s}_t)$ as the diffusion coefficient matrix which is \emph{diagonal and shared} by our \emph{LK-SDE}. Therefore, the bicycle model could be expressed as $\vb{s}_{t+1} = h(\vb{s}_t, \pi) + g_{\theta_1}(\vb{s}_t) \Delta W_t$, where the drift function $h(\vb{s}_t, \pi)$ is shown in the following Eq.~\eqref{eq:bicycle1}:

\begin{equation}
\begin{aligned}
\label{eq:bicycle1}
x_{t+1} &= x_t + \delta \cdot v_t\cos(\psi_t+\beta(u_2)),
\\
y_{t+1} &= y_t + \delta \cdot v_t\sin(\psi_t + \beta(u_2)),
\\
v_{t+1} &= v_{t} + \delta \cdot u_1
\\
\psi_{t+1} &= \psi_{t} + \delta \cdot \frac{v_t}{l_r}\sin(\beta(u_2)), \\
\beta(u_2) &= \arctan\left(\tan(u_2) \frac{l_r}{l_f + l_r}\right), \\
(u_1, u_2) &= \pi(\vb{s}_t),
\end{aligned}
\end{equation}
where the state vector $\vb{s}_t = (x_t, y_t, v_t, \psi_t)$ represents the \textit{latent representation} of lateral position, longitudinal position, velocity, and yaw angle, respectively. $\beta$ is the slip angle, and  $l_f, l_r$ are the distances between the car center and the front, and rear axle, respectively. $\delta > 0$ is a small sampling period. The control inputs correspond to the acceleration $u_1$ and front wheel steering angle $u_2$. A detailed illustration of the model is shown in Fig.~\ref{fig:bicycle_model}. 

With the bicycle model dynamics $h(\cdot)$ fixed, we implement a \textit{learnable feedback neural controller} $\pi$ to generate the control inputs as $u_1, u_2 = \pi(\vb{s}_t)$, where $\pi$ is learned from the output loss function as in Eq.~\eqref{eq:l1-smooth}.
With a learned $\pi$, the bicycle model SDE in Eq.~\eqref{eq:bicycle1} can be viewed as an autonomous system evolving with the dynamics $h(\cdot, \pi), g_{\theta_1}(\cdot)$ within the latent space representation, and providing kinematics guidance for our \emph{LK-SDE}. 



\smallskip
\smallskip
\subsubsection{Bicycle Model Guided LK-SDE}
For our \emph{LK-SDE}, the inputs for the neural network drift function $f_{\theta_0}(\cdot)$ are the latent states from the last time step, the global semantics, and local contexts. Following the transition function as shown in Eq.~\eqref{eq:neural_sde}, our \emph{LK-SDE} will generate the kinematics-aware vectors $\vb{z}$ for every time step in the latent space of the VAE:
\begin{equation}
\vb{z}_{t+1} = f_{\theta_0}(\vb{z}_t, \vb{sem}, \vb{ctx}_{t}) + g_{\theta_1}(\vb{z}_t) \Delta W_t.
\label{eq:neural_sde}
\end{equation}
To embed the kinematic knowledge from the bicycle model SDE to \emph{LK-SDE}, 
we minimize the KL divergence between the solutions (rolled latent trajectories) $(\vb{z}_t, \vb{s}_t), t \in [0, \cdots, T]$ of two SDEs (similar to the Eq. 10 in torchsde~\cite{li2020scalable} and the Appendix B in~\cite{kidger2021efficient}):
\begin{equation}
\begin{aligned}
&L_{kin} = \sum_{t=0}^{T} \mathcal{KL}(\vb{z}_t||\vb{s}_t) = \\&\sum_{t=0}^{T} \mathbb{E}_{\Delta W_t}\left[ \frac{1}{2}\norm{\biggl(g_{\theta_1}^{-1}\left(f_{\theta_0}(\vb{z}_t, \vb{sem}, \vb{ctx}_t) - h(\vb{s_t}, \pi)\right)\biggr)}^2_2\right],
\label{eq:kinematic_loss}
\end{aligned}
\end{equation}
where $T$ is the time length of the prediction and generation task. Same as the previous definition, $g_{\theta_1}$ is the diagonal and shared diffusion coefficient matrix in our \emph{LK-SDE} and bicycle model SDE. Basically, we enforce the neural drift function $f_{\theta_0}(\vb{z}_t, \vb{sem}, \vb{ctx}_t)$ to get close to the bicycle model $h(\vb{z}_t, \pi)$  for kinematics knowledge, as shown in Fig.~\ref{fig:kinematics_guidance}. In the training, we compute and minimize this loss function for every batch of samples.

Besides the kinematics loss function shown above, the \emph{LK-SDE} is also optimized by the output loss function as in Eq.~\eqref{eq:l1-smooth}, by learning directly from the dataset. Therefore, two gradient backpropagations with kinematic knowledge and data knowledge jointly improve the learning performance of \emph{LK-SDE} for more accurate, physically realistic, and controllable trajectory prediction and generation.  

\smallskip
\noindent
\textbf{Discussions:} We summarize several key points regarding our design of \emph{LK-SDE} in the following. 
\begin{myitemize}
    \item \textit{Choice of SDE:} Rather than using other dynamical models such as ordinary differential equation (ODE), we choose SDE-based latent space, as inspired by the Gaussian-distribution-based latent space in~\cite{kidger2021efficient,kingma2013auto}. This is because SDE is able to encode stochasticity in the model (the random variable at a specific timestamp of SDE follows a Gaussian distribution), which is beneficial to effectively learning real-world data with random noises. 
    \item \textit{Going beyond a single bicycle-model SDE in the latent space:} This is due to two reasons: 1) Real-world data can exhibit a wide range of behaviors and uncertainties that a single over-simplified and fixed bicycle model may not be able to account for. Introducing variability in the latent space dynamics, such as using a more flexible model or allowing parameters to change over time, can enhance the model's ability to capture diverse patterns and adapt to different scenarios. This is why we use the bicycle model to serve as a soft constraint (loss function) for the \emph{LK-SDE}. 2) The bicycle model solely considers the latent state, leading to a loss of information regarding global semantics and per-step local context, which hinders the model's learning capability. 
    \item {\textit{Choice of a learnable $\pi$ in the bicycle model for guidance:}} This design aims to introduce variability and flexibility into the latent space dynamics, addressing the limitations of the naive and over-simplified bicycle model when handling real-world data.
\end{myitemize}


\subsection{Decoder for Output and the Optimization Pipeline}
\noindent
\textbf{Decoder:} Decoder $D$ is designed as a fully-connected neural network. The decoder network $D$ takes the latent vectors $(\vb{z}_0, \cdots, \vb{z}_T)$ produced by \emph{LK-SDE} and outputs the waypoints of trajectory prediction and generation as 
\begin{displaymath}
    \hat{o}_{i:i+T} = D(\vb{z}_0, \cdots, \vb{z}_T).
\end{displaymath}

For the output waypoints, we reduce the following loss in Eq.~\eqref{eq:l1-smooth} to minimize the distance between output trajectories $\hat{o}$ and the ground truth $o$ collected in the real world: 
\begin{equation}
L_{pred}\left(o_{i},\hat {o_i}\right)= \begin{cases}0.5 (o_{i}-\hat{o_i})^{2} & \text { if }\left\|o_{i}-\hat{o_{i}}\right\|<1 \\ \left\|o_{i}-\hat{o_i}\right\|-0.5 & \text { otherwise }\end{cases}
\label{eq:l1-smooth}
\end{equation}
As a global loss function, the gradient from Eq.~\eqref{eq:l1-smooth} will back-propagate to every learnable sub-module including the GCN feature extractor, individual encoders, learnable controller in the bicycle model, \emph{LK-SDE} and the decoder. 

\noindent
\textbf{Optimization Pipeline:} Overall, the training process of our approach is shown in  Algorithm~\ref{alg:pipeline}. We optimize and balance several loss functions to regularize the latent space and generate the final trajectories. We update the feature extractor and initial state encoder by the VAE regularization loss as shown in Eq.~\eqref{eq:KL_reg}. The \emph{LK-SDE} is optimized to embed the kinematic knowledge into latent vectors by $L_{kin}$ in Eq.~\eqref{eq:kinematic_loss}, and output loss function in Eq.~\ref{eq:l1-smooth} optimizes all components.  

\smallskip
 \noindent
 \textbf{Limitations:}
The computation complexity of our approach is higher than the existing approaches because our dual SDE design in the latent space involves optimization and calculation of both kinematic and data-driven models.

\section{Experiments}
\label{sec:exp}
In this section, we conduct extensive experiments to evaluate the proposed methods. The data and training settings are introduced in Sec.~\ref{subsec:exp_set}. We demonstrate that our methods can generate physically realistic and more controllable augmented trajectories than baselines via visualized and statistical comparisons in Sec.~\ref{subsec:generation}. In Sec.~\ref{subsec:pred}, we further demonstrate the prediction accuracy of our approach and its ability to estimate unobservable variables with kinematic latent space.

\subsection{Experiment Settings}
\label{subsec:exp_set}
We train our model on the Argoverse motion forecasting dataset~\cite{chang2019argoverse} and evaluate the prediction performance on its validation set. The benchmark has more than 30K scenarios collected in Pittsburgh and Miami. Each scenario has a graph of the road map and trajectories of agents sampled at 10 Hz. In the motion generation and prediction tasks,  we use the first 2 seconds of trajectories as input and generate the subsequent 3-second trajectories.

\subsection{Physically Realistic and Controllable Trajectory Generation }
\label{subsec:generation}
As a generative model, our approach can generate diverse trajectories by tuning the latent space of our LK-SDE. 
We compare our methods with the learning-based generative model TAE~\cite{jiao2022tae} and the bicycle-model-based DKM~\cite{cui2020deep}. We measure the metrics of jerk violation rate and the Wasserstein distance of distribution of acceleration to evaluate the physical realism of generated trajectories.

The jerk is the rate of change of an object's acceleration over time (the definition is in Eq.~\eqref{eq:jerk} below) and the magnitude of jerk is commonly used to represent the smoothness of trajectories~\cite{bae2020self,scamarcio2020anti}. According to \cite{bae2020self}, the jerk threshold for discomfort presents about $0.3~m/s^3$, ranging up to $0.9~m/s^3$. In our work, we consider trajectories with jerks exceeding $0.9~m/s^3$ as violations.
\begin{equation}
\mathbf{j}(t)=\frac{\mathrm{d} \mathbf{a}(t)}{\mathrm{d} t}=\frac{\mathrm{d}^2 \mathbf{v}(t)}{\mathrm{d} t^2}=\frac{\mathrm{d}^3 \mathbf{x}(t)}{\mathrm{d} t^3}.
\label{eq:jerk}
\end{equation}

\begin{figure}[htbp]
    \centering
    \includegraphics[width=0.85\columnwidth]{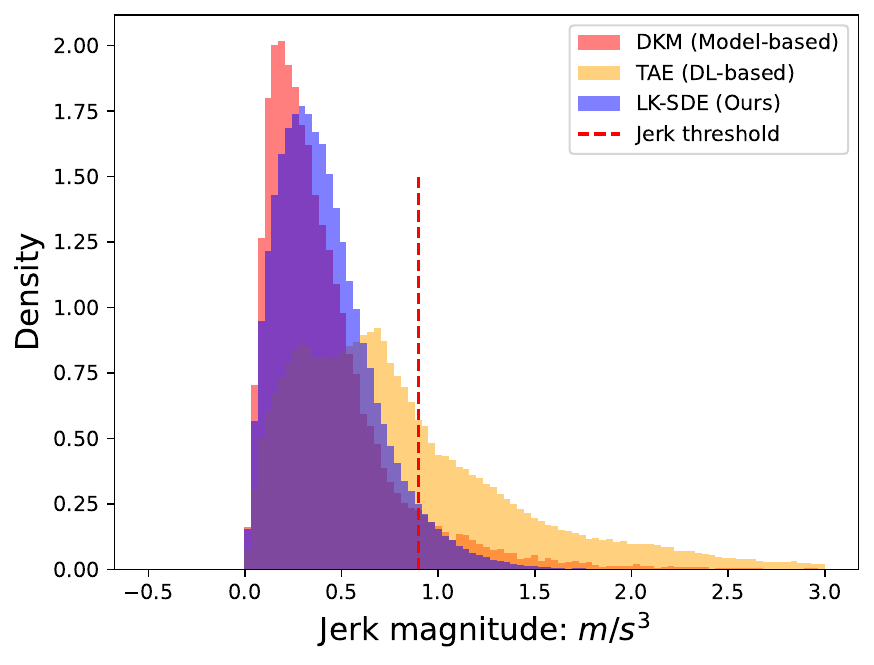}
    \caption{The distribution of the jerk magnitude of different generative methods. The red dashed line represents the discomfort threshold for jerk value. We notice that our proposed \emph{LK-SDE} can augment the smoothest trajectories.}
    \label{fig:jerk_dist}
\end{figure}

As shown in Table~\ref{tab:aug}, our \emph{LK-SDE} based model has the lowest average jerk value and lowest jerk violation rate\cite{cui2020deep}, indicating our methods can generate smoother and more realistic trajectories than DKM and TAE. Specifically, in Fig.~\ref{fig:jerk_dist}, TAE generates trajectories with sparse and unstable jerks -- about 26\% trajectories are with jerk magnitude above the discomfort threshold ($0.9~m/s^3$), showing the difficulty of physically realistic trajectory generation by pure deep learning (DL)-based methods. 
For DKM, 8.7\% of generated trajectories have higher jerk values than the threshold. \textbf{Our \emph{LK-SDE} based model generates the smoothest and physically realistic motions} with only a 5.0\% violation rate. 

We also evaluate the realism of generated motion by measuring the acceleration distribution in the thrid column of Table~\ref{tab:aug}. Previous work~\cite{liu2022statistical} shows that the forward acceleration of vehicles can be precisely described by a generalized Pareto distribution by analyzing over 100 million real-world realistic data points from more than 100,000 kilometers. We evaluate the Wasserstein distance between the motion accelerations of generated trajectories and the reference Pareto distribution. Our approach has the smallest Wasserstein distance, indicating again that \textbf{our approach can generate the most physically realistic trajectories}. 

\begin{table}[h]
\centering
\caption{Comparison between our \emph{LK-SDE} based method and baselines including TAE and DKM on the smoothness and physical realism of generated trajectories. }   

\label{tab:aug}
\resizebox{\columnwidth}{!}{
\begin{tabular}{|c|c|c|c|}
\hline
\diagbox{Model}{Metrics}   & \begin{tabular}{@{}c@{}}Average \\ Jerk \end{tabular} & \begin{tabular}{@{}c@{}}Jerk\\ Violation Rate\end{tabular} 
 &\begin{tabular}{@{}c@{}}Acc. Wasserstein \\Distance to Ref.\end{tabular} \\ \hline


TAE (DL-based)~\cite{jiao2022tae} &  $0.64$ & $26\%$ &$2.20$ \\ \hline
DKM (Model-based)~\cite{cui2020deep} &  $0.43$ & $8.7\%$ & $0.52$\\ \hline
LK-SDE (Ours)  &  $\textbf{0.40} $ & $\textbf{5.0\%}$ &$\textbf{0.45}$  \\ \hline
\end{tabular}}
\end{table}

\begin{figure*}[htbp]
\centering
\label{fig:generate_sample}

\subfigure[]{
\begin{minipage}[t]{0.28\textwidth}
\centering
\includegraphics[width=1.6in]{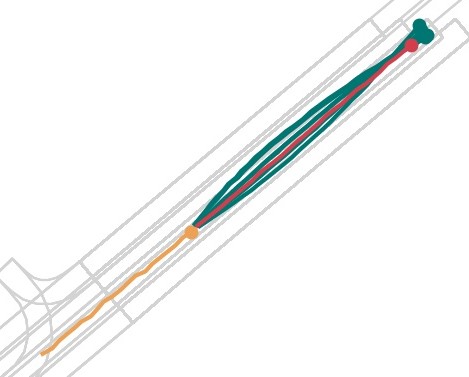}

\end{minipage}%
}%
\hfill
\subfigure[]{
\begin{minipage}[t]{0.3\textwidth}
\centering
\includegraphics[width=1.2in]{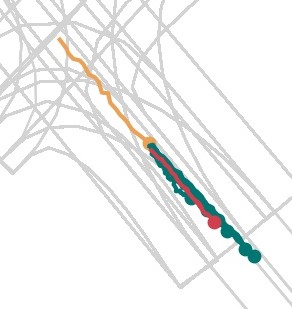}

\end{minipage}%
}%
\hfill
\subfigure[]{
\begin{minipage}[t]{0.3\textwidth}
\centering
\includegraphics[width=0.85in]{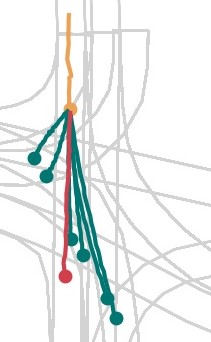}

\end{minipage}%
}%

\subfigure[]{
\begin{minipage}[t]{0.3\linewidth}
\centering

\includegraphics[width=1.0in]{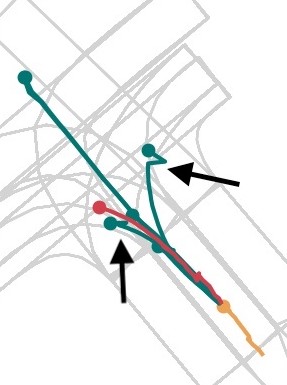}
\end{minipage}%
}%
\hfill
\subfigure[]{
\begin{minipage}[t]{0.3\linewidth}
\centering

\includegraphics[width=1.2in]{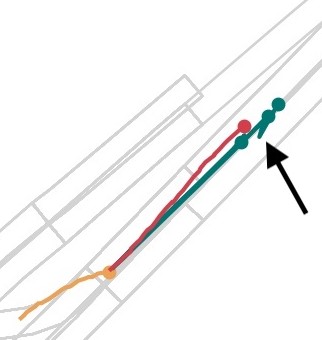}

\end{minipage}%
}%
\hfill
\subfigure[]{
\begin{minipage}[t]{0.3\linewidth}
\centering

\includegraphics[width=1.6in]{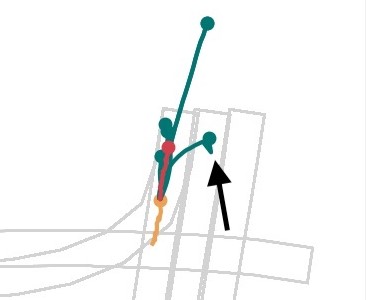}
\end{minipage}%
}%

\centering
\caption{Visualization of the augmented trajectories. The top row shows the trajectories generated by our \emph{LK-SDE} model. We manually tune the initial values in the latent space (lateral for (a), longitudinal for (b), and yaw angle for (c)).  The bottom row shows trajectories generated by the pure deep learning method TAE, by tuning its behavior latent space (lateral for (d), longitudinal for (e), random for (f)). In all the subplots, the orange lines are history trajectories, the red lines are ground truth for future motion, green lines are augmented trajectories. The black arrows point to the parts of trajectories that are obviously physically infeasible or unrealistic by TAE. The generated trajectories in (a)(b)(c) visually align with the physical knowledge from the bicycle model, showing the effectiveness of our \emph{LK-SDE} design.} 
\label{fig:sde_trajs}
\end{figure*}

Fig.~\ref{fig:sde_trajs} illustrates a few concrete examples. The top row shows that our \emph{LK-SDE} model can accurately augment realistic vehicle motions in a physically feasible and controllable manner. We tune the initial states for lateral direction, longitudinal direction, and yaw angle, respectively, which can generate the corresponding diverse trajectories. We find that many trajectories generated by the baseline TAE method (bottom row in Fig.~\ref{fig:sde_trajs}) have unrealistic and physically infeasible motions such as sharp turns and sudden offsets.

\subsection{Accurate Prediction with Physics-informed Latent Space}

\label{subsec:pred}
The accuracy of trajectory prediction is a suitable metric to measure the ability to learn the representation of the environment and generate realistic trajectories. Generally, the average performance of trajectory prediction or regression is measured by the average displacement error (ADE, defined as the average of the root mean squared error between the predicted waypoints and the ground-truth trajectory waypoints) and the final displacement error (FDE, defined as the root mean squared error between the last predicted waypoint and the last ground-truth trajectory waypoint). We compare our methods with the DL-based GRIP++~\cite{li2019grip++}, LaneGCN~\cite{liang2020learning}, and TPCN~\cite{ye2021tpcn}, domain knowledge aware generative methods TAE\cite{jiao2022tae}, and bicycle-model-based DKM\cite{cui2020deep}. The results in Table~\ref{tab:pred} show that our \emph{LK-SDE} model outperforms GRIP++, DKM, and TAE, and achieves close performance to the state-of-the-art LaneGCN and TPCN in prediction accuracy (note that LaneGCN, TPCN, and GRIP++ do not have trajectory generation capability).

\begin{table}[h]
\centering
\caption{Trajectory prediction comparison between our \emph{LK-SDE} based method and baselines. 
}

\label{tab:pred}
\resizebox{0.6\columnwidth}{!}{
\begin{tabular}{|c|c|c|c|}
\hline
\diagbox{Model}{Metrics}   & ADE & FDE \\ \hline

GRIP++\cite{li2019grip++}  & $1.77$  &$3.91$ \\ \hline 
LaneGCN\cite{liang2020learning}  & $1.35$  &$2.96$ \\ \hline
TPCN\cite{ye2021tpcn} & $1.34$  &$2.95$ \\ \hline
TAE \cite{jiao2022tae} &  $1.42$ &$3.08$ \\ \hline
DKM \cite{cui2020deep} &  $1.46 $ &$ 3.14$ \\ \hline
LK-SDE (Ours)  &  $ 1.39$ &$2.98 $ \\ \hline
\end{tabular}}
\vspace{-12pt}
\end{table}

\begin{figure}[htbp]
    \centering
    \includegraphics[width=0.8\columnwidth]{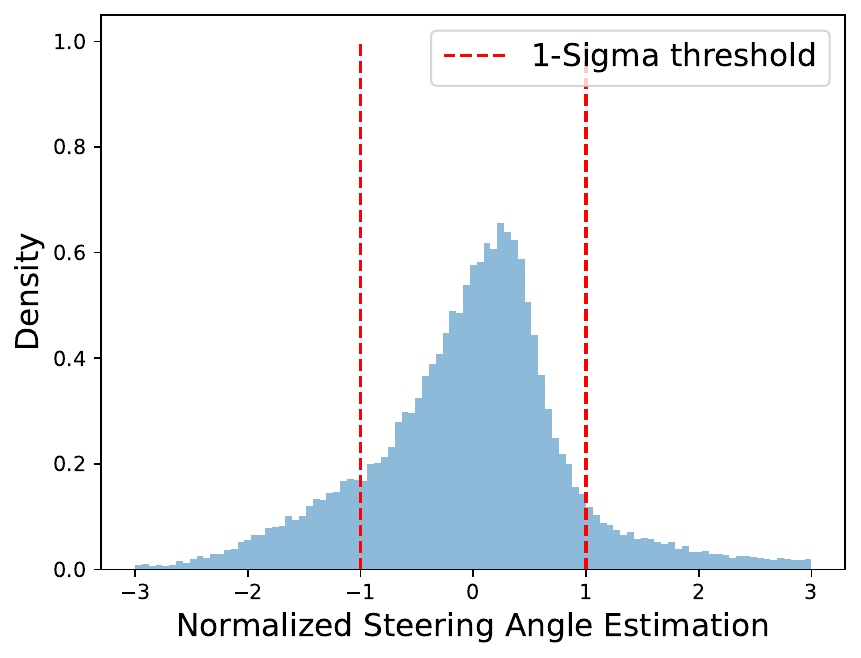}
    \caption{The distribution of the normalized steering angles estimated in the latent space. The red dashed line represents the one-sigma threshold. Our kinematics-aware method can estimate unobservable variables such as steering angle, which is useful for understanding target vehicles' intentions and behavior patterns. }
    \label{fig:steering_dist}
\end{figure}

Moreover, our kinematics-aware latent space can provide estimations on some variables that are crucial for understanding the target vehicle's behavior but cannot be observed directly by perception or prediction modules. In our design, the bicycle-model-based LK-SDE can estimate the normalized steering angles ($u2$) and slip angles ($\beta$) without explicit training on ground truth data. Fig.~\ref{fig:steering_dist} shows the distribution of normalized steering angles on the Argoverse dataset, estimated by the kinematics-aware latent space. The normalized steering angle and its distribution can help us discern the steering directions as well as the sharpness of the turns. For instance, when the estimated steering angle of a vehicle is larger than a threshold (such as the one-sigma range in Fig.~\ref{fig:steering_dist}),  it suggests the possibility of an aggressive turning maneuver. 

Combining the results from trajectory generation and augmentation, we can conclude that our latent kinematics-aware SDE can learn the representation of trajectories and environments more effectively than the model-based methods and generate more physically realistic and controllable motions than the DL-based generative models. In addition to the average accuracy, our methods can also \textbf{give detailed and accurate kinematics prediction (e.g., steering angle) along with the waypoints}, which provide more explainable information for safety-critical decision-making.


\section{Conclusion}
In this work, we propose a vehicle motion generator with the latent kinematics-aware stochastic differential equation (\emph{LK-SDE}). We embed the physics knowledge into the latent space of a VAE by the dual SDE design. The method can bridge the high-dimensional features from the environment and the low-dimensional kinematics to generate fine-grained,  physically realistic, and controllable trajectories, and to provide accurate prediction of unobservable physical variables. 

\label{sec:con}




\bibliographystyle{IEEEtran}
\bibliography{reference}

\end{document}